\documentclass[letterpaper, 10pt, conference]{ieeeconf}
\IEEEoverridecommandlockouts
\usepackage[utf8]{inputenc} 
\usepackage[T1]{fontenc}    
\usepackage{hyperref}       
\usepackage{url}            
\usepackage{booktabs}       
\usepackage{amsfonts}       
\usepackage{nicefrac}       
\usepackage{microtype}      
\usepackage{graphicx}
\usepackage{grffile}
\usepackage{amsmath}
\usepackage{algpseudocode}
\usepackage{subcaption}
\usepackage[subrefformat=parens]{subcaption}
\usepackage{subcaption}
\usepackage{multirow}
\usepackage{multicol}
\usepackage{bm}
\usepackage{color}

\newcommand{\argmin}{\mathop{\rm arg~min}\limits}
\newcommand{\Tref}[1]{Table~\ref{#1}}
\newcommand{\eref}[1]{Eq.~(\ref{#1})}

\newcommand{\fref}[1]{Fig.~\ref{#1}}
\newcommand{\Fref}[1]{Figure~\ref{#1}}
\newcommand{\sref}[1]{Sec.~\ref{#1}}

\title{\LARGE \bf
	Efficient Exploration in Constrained Environments \\with Goal-Oriented Reference Path}

\author{Kei Ota$^{1,2}$, Yoko Sasaki$^{2}$, Devesh K. Jha$^{3}$, Yusuke Yoshiyasu$^{2}$, and Asako Kanezaki$^{2}$
	\thanks{$^{1}$Kei Ota is with Information Technology R\&D Center, Mitsubishi Electric Corporation, Japan.
		{\tt\small Ota.Kei@ds.MitsubishiElectric.co.jp}}%
	\thanks{$^{2}$Kei Ota, Yoko Sasaki, Yusuke Yoshiyasu and Asako Kanezaki are with National Institute of Advanced Industrial Science and Technology (AIST), Japan.
		{\tt\small y-sasaki, yusuke-yoshiyasu, kanezaki.asako@aist.go.jp}}%
	\thanks{$^{3}$Devesh K. Jha is with Mitsubishi Electric Research Labs, Cambridge, MA, USA.
		{\tt\small jha@merl.com}}%
}

\begin{document}
    \twocolumn[{%
    \renewcommand\twocolumn[1][]{#1}%
    \maketitle
    \begin{center}
        \centering
        \includegraphics[width=.9\textwidth]{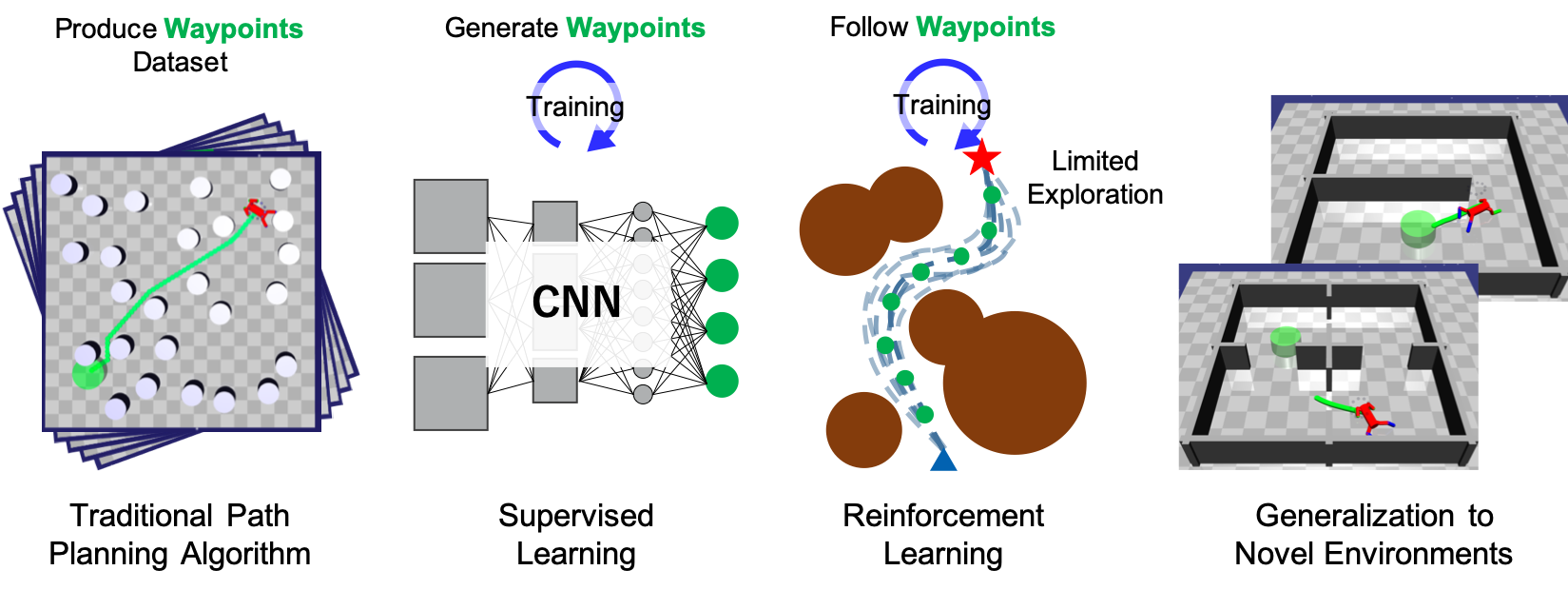}
        \captionof{figure}{
            We decouple planning and control by combining traditional path planning algorithms, supervised learning (SL) and reinforcement learning (RL) algorithms in a synergistic way.
            First, a path planning algorithm such as A* search generates a set of pairs of aerial images of environments and waypoints, and then a convolutional neural network (CNN) learns the general rule of planning by SL and generates  waypoints for the agent to follow. The agent learns to follow arbitrary waypoints while using path-conditioned RL and thus, resulting in efficient exploration. We show that our trained agent can achieve good sample efficiency as well as generalization to novel environments.
        }
        \label{fig:proposed_method}
    \end{center}%
    }]
    \footnotetext[1]{Kei Ota is with Information Technology R\&D Center, Mitsubishi Electric Corporation, Japan.
   {\tt\small Ota.Kei@ds.MitsubishiElectric.co.jp}}%
  \footnotetext[2]{Kei Ota, Yoko Sasaki, Yusuke Yoshiyasu and Asako Kanezaki are with National Institute of Advanced Industrial Science and Technology (AIST), Japan.
   {\tt\small y-sasaki, yusuke-yoshiyasu, kanezaki.asako@aist.go.jp}}%
  \footnotetext[3]{Devesh K. Jha is with Mitsubishi Electric Research Labs, Cambridge, MA, USA.
   {\tt\small jha@merl.com}}%

    \thispagestyle{empty}
    \pagestyle{empty}
    \begin{abstract}
  In this paper, we consider the problem of building learning agents that can efficiently learn to navigate in constrained environments. The main goal is to design agents that can efficiently learn to understand and generalize to different environments using high-dimensional inputs (a 2D map), while following feasible paths that avoid obstacles in obstacle-cluttered environment.
  To achieve this, we make use of traditional path planning algorithms, supervised learning, and reinforcement learning algorithms in a synergistic way. The key idea is to decouple the navigation problem into planning and control, the former of which is achieved by supervised learning whereas the latter is done by reinforcement learning. Specifically, we train a deep convolutional network that can predict collision-free paths based on a map of the environment-- this is then used by an reinforcement learning algorithm to learn to closely follow the path.  
 This allows the trained agent to achieve good generalization while learning faster. We test our proposed method in the recently proposed \textit{Safety Gym} suite that allows testing of safety-constraints during training of learning agents. We compare our proposed method with existing work and show that our method consistently improves the sample efficiency and generalization capability to novel environments.
\end{abstract}

    \section{Introduction}


Designing RL agents that can learn complex, safe behavior in constrained environments for navigation has been getting a lot of attention recently~\cite{Ray2019safetygym}. This has been mainly driven by the race to achieve artificial general intelligence where AI agents can achieve human-like performance~\cite{lake2015human}. It is desirable that the RL agents can generalize to novel environments while achieving optimal performance. Motivated by this problem, in this paper, we consider the problem of building learning agents that can efficiently learn to navigate in constrained environments. In this paper, we present a sample-efficient way of designing agents that can learn to generalize to different environments by combining classical planning , supervised learning and reinforcement learning (see \fref{fig:proposed_method} for pictorial explanation of the proposed method).

Motion planning is central to robotics and thus has been extensively studied~\cite{Stentz1994,Arambula2004,lavalle2006planning}. 
Most of these techniques explicitly considering the geometry of obstacles as well as the robot and find feasible paths ensuring collision avoidance while guaranteeing performance. However, most of these methods handle only kinematic and geometric planning and do not necessarily provide control trajectories that the robot can use to follow the planned path~\cite{lavalle2006planning}. Furthermore, these methods may have difficulties when the planning space becomes larger or the workspace of the robot changes dynamically.

Deep Learning has been to achieve super-human performance in the fields of supervised image recognition~\cite{resnet,densenet} as well as when used with RL algorithms for learning games~\cite{mnih2015human}. Deep RL is built on top of recent progresses of Deep Neural Networks (DNNs), which enables it to work even in a large state space such as 2D images, 3D point clouds, etc., by exploiting the high representation capability of DNNs. Also, it enables an agent to acquire complex behaviours considering physical constraints by neatly designing reward functions.
Deep RL, however, needs huge amounts of interaction with environments, and could easily overfit to perform optimally in a particular environment. Deep RL has been recently applied to real-time motion planning tasks~\cite{everett2018motion, lotjens2019safe}. 

Our goal is to efficiently train AI agents that can learn to navigate constrained environments while ensuring safety of the robot. Our proposed method is explained pictorially in \fref{fig:proposed_method}. We make use of traditional path planning algorithms, supervised learning (SL) and RL algorithms. We use these three components in a synergistic way to train agents efficiently which can learn to generate feasible trajectories while generalizing to novel environments. Our main idea is to decouple planning and control for an agent for navigation in obstacle-cluttered environment. We use SL to generate waypoints (or subgoals) and the agent then uses RL to learn to follow any sequence of waypoints by using path-conditioned RL. This allows us to generalize to novel environments as the SL can generate waypoints for arbitrary environments after training and the agent can follow any arbitrary trajectory after training with RL. Furthermore, the goal-oriented reference path encourages the RL agent to explore limited regions and thus helps improve sample efficiency. We test our proposed method in the recently proposed \textit{Safety Gym}~\cite{Ray2019safetygym} suite that allows testing of safety-constraints during training of learning agents. We compare our proposed method with existing work and show that our method consistently improves the sample efficiency and generalization capability.

    \section{Related Work}
Deep RL has recently seen tremendous success in learning complex policies for several complex sequential tasks and has attracted huge attention by often achieving superhuman performance~\cite{mnih2015human,silver2017go}. While this shows that deep RL can allow agents to learn complex policies~\cite{schulman2015trust, lillicrap2015continuous}, it is also clear that these algorithms require impractical amount of data to learn behavior which are oftentimes very intuitively simple. This has driven a lot of research to allow agents to learn efficiently by allowing better exploration or improving the convergence rates of RL algorithms. Our work falls in the category of allowing better exploration strategies for the agents during learning. Under this category, our work draws similarity with existing work on RL with reference trajectory and also broadly to hierarchical RL~\cite{kulkarni2016hierarchical}.  



The combination of reference paths and RL has been widely researched \cite{faust2018prm,Thomas_2018,chiang2019learning,ota2019trajectory}.
In \cite{faust2018prm}, Probabilistic Roadmaps (PRM)~\cite{kavralu1996probabilistic} is used to find reference paths, and RL is used to point-to-point navigation as a local planner for PRM.
In contrast, \cite{chiang2019learning} uses RL to learn both point-to-point and path-following navigation, which however requires a huge training time because it involves hyper-parameter tuning to improve performance.
In \cite{Thomas_2018}, a model-based RL agent is learned for autonomous robotic assembly by exploiting the prior knowledge in the form of CAD data, using a Guided Policy Search (GPS)~\cite{levine2013guided} approach to learn a trajectory-tracking controller. However, the GPS algorithm still produces very local policies and cannot generalize to changing environments.
The closest work to ours is \cite{ota2019trajectory}, which learns an RL agent that optimizes trajectory for 6-DoF manipulator arm. They produce a reference path by Rapidly-exploring Random Trees (RRT)~\cite{lavalle2006planning} and those points are then used to compute informative reward to reach goal.
The important difference is that~\cite{ota2019trajectory} assume obstacles and desired location do not change during the whole training/test phase. Our approach, however, can deal with changes in environments by conditioning an RL agent with reference path which is also generated based on an observation of current environment.


Our combination of SL and RL can be regarded as HRL, where a higher-level policy generates a rough plan to accomplish a task and lower-level agent generates primitive actions to control a robot~\cite{parr1998reinforcement,sutton1999hrl}. This decomposition is often used to either simplify a task or reduce the sample complexity for learning~\cite{stolle2002learning}.
For example in~\cite{bischoff2013hierarchical}, the authors train both discrete and continuous RL for movement planning and execution. However, this was studied only for grid world. 

Similar to our work, SL is used to learn a path generator in \cite{bansal2019-lb-wayptnav}.
The navigation is however limited to small environments because it uses first-person-view images captured by a single on-board camera.
In order to deal with a navigation task in a large environment, we use aerial view maps as in \cite{tamar2017vin,kanezaki2018goselo,bansal2019chauffeurnet}.
Another related work~\cite{qureshi2018deeply,qureshi2019motion} proposes neural network-based motion planning network.
They learn to produce feasible paths directly from a point cloud measurement, and shows the high generalization capability.
However, their approach considers only path generation, which is similar to the SL part of our work, and requires to track the generated path using some kind of lower-controller, and this could be difficult for a robot that has high dimensional state and action space.
Our approach solves the full problem as the waypoints generator is responsible for generalizing to unseen environments, and RL can generate primitive actions that realizes to track the path even in a high dimensional dynamical system.

    \section{Background \label{sec:background}}
In this section, we provide some background for algorithms and methods used in our paper.
We describe the Soft-Actor Critic (SAC) algorithm \cite{haarnoja2018soft} and a state representation \cite{kanezaki2018goselo} which are used in our proposed method to achieve efficiency and generalization capability.

\subsection{Reinforcement Learning}
We consider the standard RL setting that consists of an agent interacting with a stochastic environment.
An environment consists of a set of states $\mathcal{S}$, a set of actions $\mathcal{A}$, a distribution of initial states $p(s_0)$, a reward function $r : \mathcal{S} \times \mathcal{A}\rightarrow \mathbb{R}$, transition probabilities $p(s_{t+1}|s_t, a_t) : \mathcal{S} \times \mathcal{A}\rightarrow \mathcal{S}$, and a discount factor $\gamma \in [0,1]$.

An episode starts with an initial observation $s_0$ sampled from $p(s_0)$.
At each time step $t$, the agent observes an observation $s_t$ and chooses an action $a_t$ according to a policy $\pi(a_t|s_t)$, which is a mapping from observations to actions: $\pi : \mathcal{S} \rightarrow \mathcal{A}$.
Then, the agent obtains a reward $r_t = r(s_t, a_t)$, and the next state $s_{t+1}$ is sampled from $p(s_{t+1}|s_t,a_t)$.
The goal of the agent is to maximize the expected discounted sum of rewards $J = \mathbb{E}_\pi[ \sum_{t=0}^{\infty}\gamma^t r(s_t,a_t) ]$.
The quality of the agent's action $a_t$ when receiving an observation $s_t$ can be measured by a $Q$ function $Q(s_t, a_t) = \mathbb{E}_\pi[ J | s_t, a_t ]$.

\subsection{Soft Actor Critic \label{subsec:sac}}
We use Soft Actor Critic (SAC)~\cite{haarnoja2018soft}, which is an off-policy actor-critic algorithm for our experiments.
In actor-critic algorithms, the critic learns the action-value function $Q_{\pi}(o_t, a_t)$, while the actor learns a policy $\pi(a_t | o_t)$ to select frequently $a_t$ that has a high value in $Q_{\pi}$.
The policy of SAC gives the distribution of actions $\pi(a_t | o_t)$, and the action is determined stochastically during training.
SAC favors the policy that maximizes not only the expected sum of rewards, but also the expected entropy of the distribution ${\cal H} (\pi (a_t | o_t))$. Consequently, its objective function is defined as:
\begin{equation}
    \tilde{J}(\pi) = \sum_{t=0}^{T} \mathbb{E}_{(o_t, a_t) \sim \rho_{\pi}(o_t, a_t)} [r_{t+1} + {\cal H}(\pi(\cdot | o_t))]
\end{equation}
where $T$ is the final time step of an episode, and $\rho_{\pi}(o_t, a_t)$ represents the distribution of observation-action pairs given policy $\pi$. 


\subsection{GOSELO}
    We borrow the generalization capability from GOSELO (Goal-directed Obstacle and SElf-LOcation), which provides a representation of an image with multiple channels for the input to a CNN~\cite{kanezaki2018goselo}.
    GOSELO is a combination of two bird's-eye view maps: a map of obstacles observed by the agent's laser scanner and a map of self-location history from the beginning to the current state.
    The pixels of the former map have binary values, where $1$ indicates a pixel is occupied by an obstacle, and $0$ indicates a free or invisible space. Each pixel of the latter map has an integer value that represents how many times the agent has visited a location.
    These maps are transformed in the manner as first, rotate and translate maps so that the goal $G$ is located directly above the current location $P$ and the center point $M$ is at the center of the entire image. Next, the rotated image is cropped as a square of size $(L + 4) \times (L + 4)$ centered at $M$, where $L$ denotes the number of the pixels on the line $GP$. The rotated image is then also cropped squares of size $\alpha L \times \alpha L$ and merge them as additional channels. Here, we use two additional channels with $\alpha = (2, 4)$ which is experimentally chosen considering the field size of the environment we use. Finally, those images are concatenated as six channels (i.e., an image of size $W \times H \times 6$) as the input of a CNN, which consists of three channels from an obstacle map and the three channels from a self-location history.
    The motivation of GOSELO is to make the input environment representation irrelevant to the specific locations of start and goal as well as the shape of obstacles, so as to achieve \textit{general} navigation policy.
    In \cite{kanezaki2018goselo}, the policy is trained in a supervised learning manner where the training data are generated using A* search~\cite{hart1968astar}.

    \section{Proposed Method}
In this section, we present details of the proposed method. Specifically, we describe our method of training an agent using SL followed by RL. Figure~\ref{fig:architecture} shows a schematic of the proposed method used to train agents in the paper.

\subsection{Waypoints Generator}\label{subsec:waypoint_gen}
Using high-dimensional inputs (images) for training RL agents generally requires huge number of samples for learning meaningful policies as they require bigger networks and thus, more data for convergence~\cite{lesort2018srlreview}. However, for navigation tasks, generating an optimal path (i.e., the sequence of waypoints) is easy by using traditional path planning algorithms, and imitating those optimal paths using DNNs with SL fashion is also easy. We use this idea to first train a module for the agent that can predict a rough path using a map of the environment.

In order to train such an optimal waypoints generator, we first generate a dataset that consists of pairs of GOSELO-style inputs $o_k$ and optimal paths $z_k$. The optimal paths are generated by A* algorithm~\cite{hart1968astar} from a random start, goal, and obstacles location. Then, the waypoints generator learns to produce the waypoints by minimizing the following objective:
\begin{equation}
	L_{\rm waypoints} = \frac{1}{K} \sum_{k=1}^{K} \| g(o_k) - z_k \|^2_2 \label{eq:sl}.
\end{equation}

The waypoints generator, $g$ in \eref{eq:sl}, consists of a convolutional neural network (CNN) that takes a GOSELO-style input image and generates a short horizon path that consists of $10$ waypoints, each of which has $2$ dimensional relative position to the current location of the agent. Thus, the SL module provides the agent a rough, collision-free path that it can follow to reach the desired goal. Next we describe a path-conditioned RL framework where the agent learns to closely follow these waypoints in a collision-free fashion.

\subsection{Path Conditioned Reinforcement Learning}\label{subsec:path_conditioned_rl}
\begin{figure*}
	\centering
	\includegraphics[width=\linewidth]{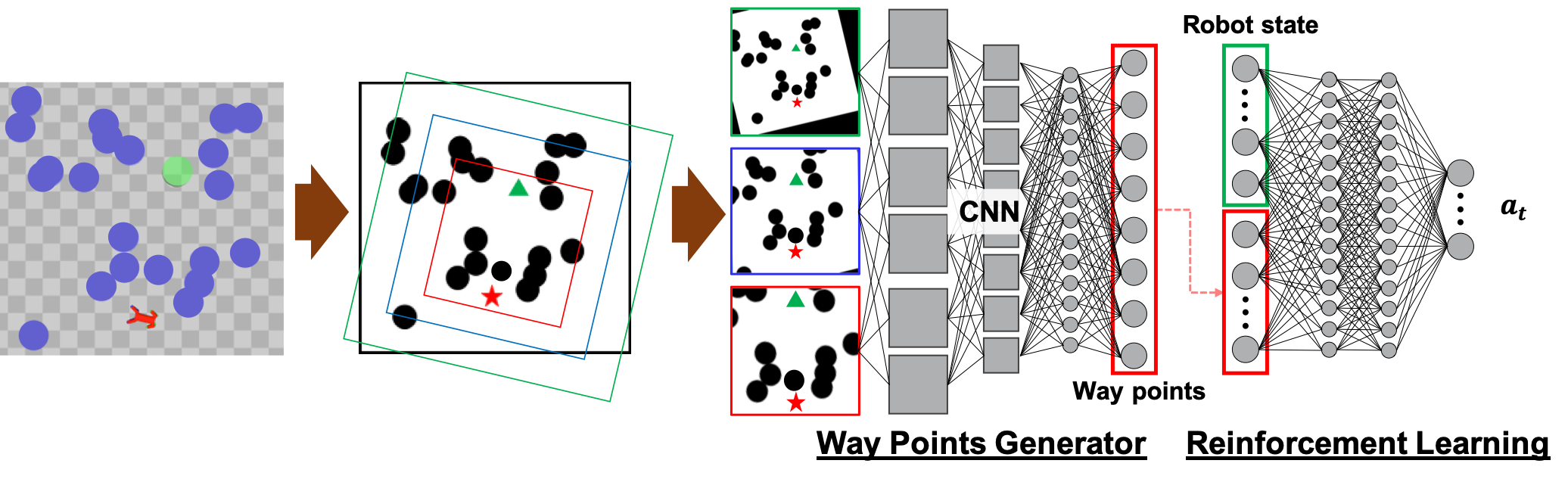}
	\caption{Architecture of the proposed method. The aerial view of a field is translated into a binary map and then converted to a GOSELO style image. Waypoints generator takes the image and outputs the waypoints that roughly guides the agent to the desired goal. The input to the RL agent is concatenation of the waypoints and an internal state of each robot, and outputs a low-level action to control the robot.}
	\label{fig:architecture}
\end{figure*}

We consider the standard RL problem described in \sref{sec:background} with the waypoints $z \in \mathcal{Z}$ that lead an RL agent to the goal location.
\Fref{fig:architecture} shows the architecture of the proposed approach.  As mentioned earlier, the waypoints generator takes a GOSELO style image as an input and generates a sequence of waypoints. The low-level controller (an RL agent) then takes the sequence of the waypoints and  internal robot state as an input, and generates low-level action to closely track the waypoints.
The waypoints $\bm{z}$ can be considered as an intermediate representation of a state, because it is acquired by learning a map of GOSELO-style image to the waypoints. We propose to exploit this information to improve the performance of RL agents by combining it with a conventional reward function as $r : \mathcal{S} \times \mathcal{A} \times \mathcal{Z} \rightarrow \mathbb{R}$, and inputs to an RL agent as $o : \mathcal{S} \times \mathcal{Z} \rightarrow \mathcal{O}$, where $\mathcal{O}$ is the extended state space for the agent.
Therefore, the reward function can be written as
\begin{equation}
r(s_t, a_t, z_t) = f(s_t, a_t) + h(s_t, a_t, z_t),
\end{equation}
where $f(s_t, a_t)$ is the reward that originates from an original RL setting. We obtain $h(s_t, a_t, z_t)$ using the waypoints.
%
We specifically define the reward function $h(s_t, a_t, z_t)$ as:
\begin{equation}
    h(s_t, a_t, z_t) = w_1 d_{\rm path} + w_2 n_{\rm progress},
\label{eq:reward_way_points}
\end{equation}
where $d_{\rm path}$ is the distance to the reference path and $n_{\rm progress}$ is the progress along the path.
The first term limits exploration area by penalizing the distance to the waypoints, whereas the second term encourages the agent to move along the waypoints towards the goal location.

\begin{figure}[tpb]
	\centering
	\includegraphics[width=0.7\columnwidth]{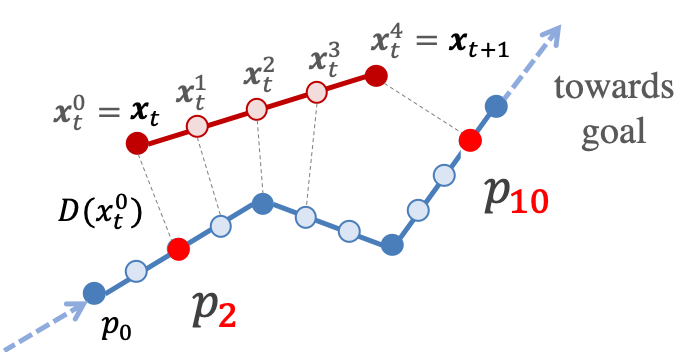}
	\caption{
			Path division for calculating rewards.
			The blue line is the part of waypoints that leads to a goal location and the red line is the agent's path.
			$\bm{x}_t$ is the position of the agent at time step $t$ and $\bm{p}_j$ is the index of the divided waypoints.
			The dashed lines indicate the correspondence to the closest neighbor.}
	\label{fig:path}
\end{figure}

In order to calculate $d_{\rm path}$ and $n_{\rm progress}$, we divide the waypoints and agent's path at regular intervals, as shown in \fref{fig:path}.
By dividing the path, we obtain the sub-sampled vertices $\bm{p}_0, \bm{p}_1, \cdots, \bm{p}_{N_p-1}$ for the waypoints,
and $\bm{x}^0_t, \bm{x}^1_t, \cdots, \bm{x}^{N_t-1}_t$ for the agent's path,
where $N_p'$ and $N_t$ are the numbers of vertices in each divided path.
We can then define the distance to the given path as
$d_\text{path} = \max\limits_{i} D(\bm{x}_t^i)$,
where $D(\bm{x})$ is the distance to the path calculated as $\min\limits_{i}\|\bm{x} - \bm{p}_i\|$.
We can also observe the progress along the path as
$n_{\rm progress} = {\rm NNI}(\bm{x}_{t+1}) - {\rm NNI}(\bm{x}_{t})$,
where ${\rm NNI}(\bm{x})$ is the vertex index of the nearest neighbor to $\bm{x}$,
i.e., ${\rm NNI}(\bm{x}) = \argmin_{i}\|\bm{x} - \bm{p}_i\|$.
In the case of \fref{fig:path}, $n_{\rm progress} = 10 - 2 = 8$.

This reward function is then used by an RL algorithm to train a policy to follow the sequence. An important point to note here is that the RL agent learns path-conditional policy, i.e., a path-tracking policy. This allows flexibility in achieving generalization as the RL is not trained specifically for a goal. As a result, as long as the waypoints generator provides a feasible path to the RL module, the agent can follow it to reach the desired goal. In the next section, we describe how we update the waypoints during execution to ensure collision-free motion of the agent.

\subsection{Online Replanning}\label{subsec:online-replanning}
In the previous two sections, we described our proposed method for generating feasible waypoints and then training the RL agent to follow a sequence of waypoints. As we discussed in \sref{subsec:waypoint_gen}, the waypoints generator outputs only $10$ waypoints every-time it is queried. However, since the waypoints are not generated considering the dynamics of the robot, the robot doesn't exactly follow these waypoints and might start deviating from it after a short horizon. Consequently, we keep updating the waypoints after a certain time window. This is done by updating the state of the robot (and the environment in the presence of moving obstacles) in the GOSELO-style image and then generate new waypoints. The robot then tries to follow the new sequence of waypoints. This is then repeated till the robot reaches the desired goal. Thus the agent needs to decide when it should update its plan. Ideally, updating its plan after every step will result in optimal behavior but may be computationally heavy as computation of waypoints requires the forward propagation of relatively computationally heavy CNN architecture.
On the other hand if the update frequency is too small, the agent might not be able to move efficiently.
Therefore, the update frequency should be defined by a trade-off between optimal behaviour and the computational cost.
To solve this trade-off, we propose a rule-based online replanning strategy, where the agent asks for new waypoints depending on whether 1) the distance of the agent and the closest way point $d_{\rm path}$ is larger than a threshold $d_{\rm update}$, or 2) the closest index of waypoints ${\rm NNI}(x_t)$ is larger than a half of the number of waypoints, where $d_{\rm path}$ and ${\rm NNI}(x_t)$ are defined in \sref{subsec:path_conditioned_rl}. The threshold in condition (1) is a hyperparameter that needs to be tuned.

    \section{Experimental Settings}\label{sec:experiments}

This section briefly describes the experimental settings including the environments and tasks used in the paper to validate the proposed method.

\subsection{Environment}
We evaluate our method on custom environment, which is based on Safety Gym~\cite{Ray2019safetygym}.
Here we describe the details of the environment.

\newcommand{\point}{\textit{point}}
\newcommand{\car}{\textit{car}}
\newcommand{\doggo}{\textit{doggo}}

\subsubsection{States}
The states of an agent includes the standard robot sensors (accelerometer, gyroscope, magnetometer, velocimeter, and robot-centric lidar observation of 10 directions exclusive and exhaustive for a full 360 view) and robot specific sensors (such as joint positions and velocities) as $\bm{s}_{\rm robot}$, and $10$ waypoints $\bm{z}\in\mathbb{R}^{2\times10}$ generated by the way-points generator that guides the agent to the goal location as described in~\sref{subsec:waypoint_gen}.

\subsubsection{Actions}
The actions of the agent is to control each robot to achieve a desired task.
The robot we use for our experiments include three types: \point, \car, and \doggo.
\textit{point} is a simple robot constrained to the 2D-plane, with one actuator for turning and another for moving forward/backwards.
\textit{car} has two independently-driven parallel wheels and a free rolling rear wheel.
\textit{doggo} is a quadrupedal robot with bilateral symmetry. Each of the four legs has two controls at the hip, for azimuth and elevation relative to the torso, and one in the knee, controlling angle.
For more details, please refer~\cite{Ray2019safetygym}.
Table~\ref{tab:mdp} summarizes the number of state and action dimensions for each robot type.

\begin{table}[tbp]
    \centering
    \caption{Number of dimensions of states and actions for each robot.}
    \begin{tabular}{ccccccc} \toprule
        Robot type & States & Actions \\ \midrule
        \point & $42$ & $2$ \\
        \car & $54$ & $2$ \\
        \doggo & $78$ & $12$ \\ \bottomrule
    \end{tabular}
    \label{tab:mdp}
\end{table}

\subsubsection{Rewards}\label{subsubsec:rewards}
    As described in \sref{subsec:path_conditioned_rl}, the reward function of our path-conditioned RL consists of conventional reward function $f(s_t, a_t)$, and $h(s_t, a_t, z_t)$ which is calculated from the waypoints.
    Since the $h(s_t, a_t, z_t)$ is already described, here we define the reward term for original RL setting referring to \cite{Ray2019safetygym} as:
	\begin{equation}
	    f(s_t, a_t) = w_3 \mathbb{I}_{\rm collision} + w_4 \mathbb{I}_{\rm goal},
    	\label{eq:reward_rl}
	\end{equation}
	where $\mathbb{I}_{\rm collision}$ and $\mathbb{I}_{\rm goal}$ are indicator functions of whether collision between the agent and the obstacles occurs, and whether the agent reaches goal location.

\subsubsection{Terminal conditions}
    An episode terminates with following three conditions: the location of the agent $\bm{x}_t$ is sufficiently close to the goal location, the agent explores too far from the goal location, or the number of steps of an episode is over a specified threshold.

    \section{Experimental Results}\label{sec:results}
This section presents results from experiments designed to answer the following questions:
\begin{itemize}
	\item Can we learn to solve the desired task of reaching specified goal while avoiding collision with randomly-placed obstacles?
	\item Does the combination of waypoints generator and RL-based controller improve performance on RL?
	\item Does the waypoints generator helps low-level controller generalize to novel environments?
\end{itemize} 

\subsection{Performance of Waypoints Generator}\label{subsec:results_way_points_gen}
    We first demonstrate the effectiveness of the waypoints generator which generates the waypoints for the RL agent. 
    In order to evaluate the quality of paths, we move the agent to the closest waypoint for each step, and evaluate the average number of times of collision over $1000$ episodes.

    Table~\ref{tab:results_way_points_gen} shows the results for average number of collisions.
    It suggests that the quality of waypoints improves when we add training data, and the number of collisions in an episode saturates around $1.00$.
    The inaccuracy of the waypoints generator will be taken care of by the RL agent by minimizing the collision penalty defined in \eref{eq:reward_rl}.

	\begin{table}[tbp]
		\caption{Average number of collision with obstacles in an episode.}
		\label{tab:results_way_points_gen}
		\begin{center}
			\begin{tabular}{ccccc} \toprule
			     Number of training data & $1$K & $10$K & $50$K & $100$K \\ \midrule
			     Number of collisions in an episode & $3.05$ & $1.57$ & $1.15$ & $0.96$ \\ \bottomrule
			\end{tabular}
		\end{center}
	\end{table}

\subsection{Sample Efficiency\label{subsec:vs_baseline}}
    \begin{figure*}[t]
    	\begin{minipage}[]{0.43\linewidth}
    		\centering
    		\includegraphics[width=\columnwidth]{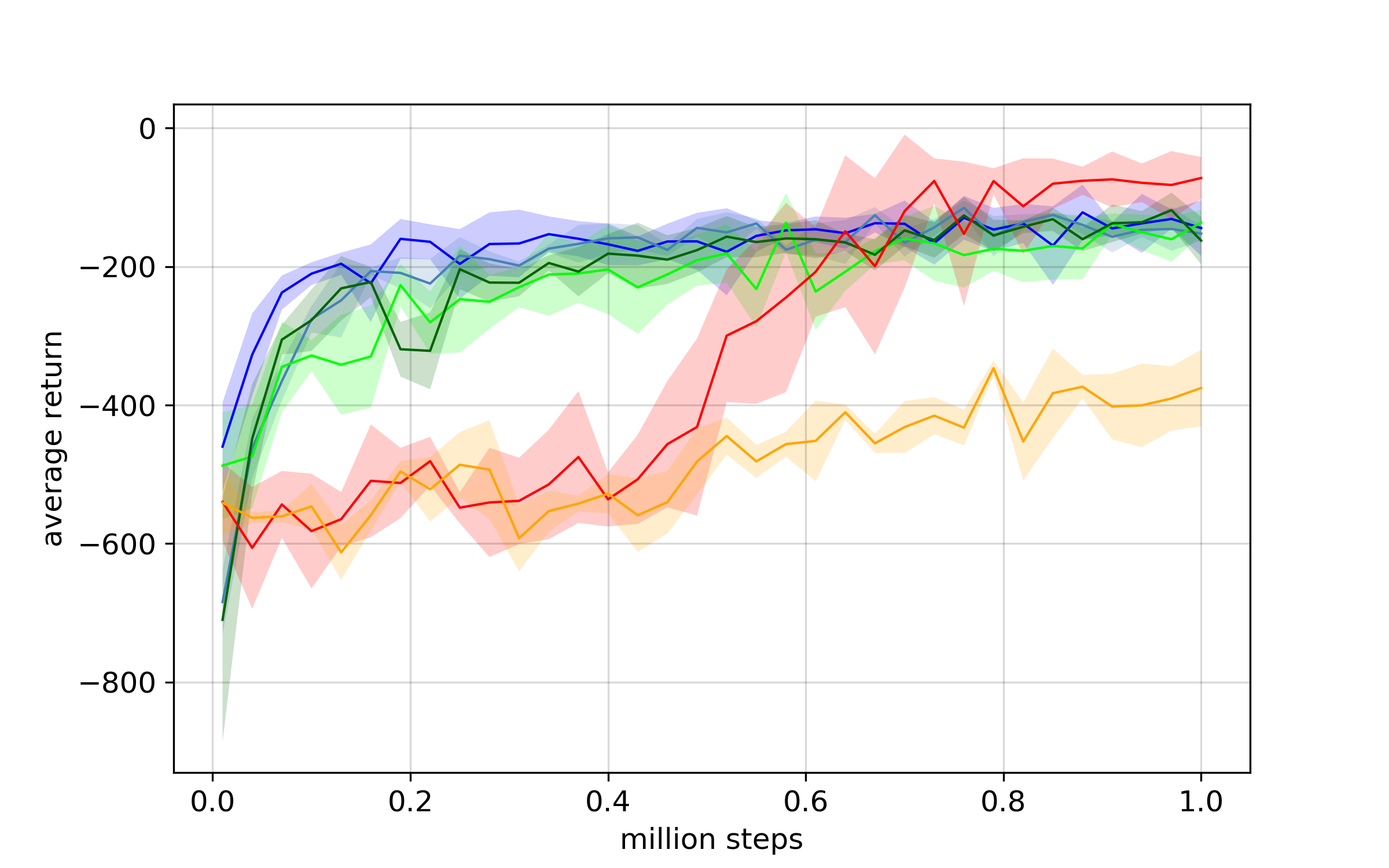}
    		\subcaption{Average return (higher is better)}\label{fig:results_return}
    	\end{minipage}
    	\begin{minipage}[]{0.43\linewidth}
    		\centering
    		\includegraphics[width=\columnwidth]{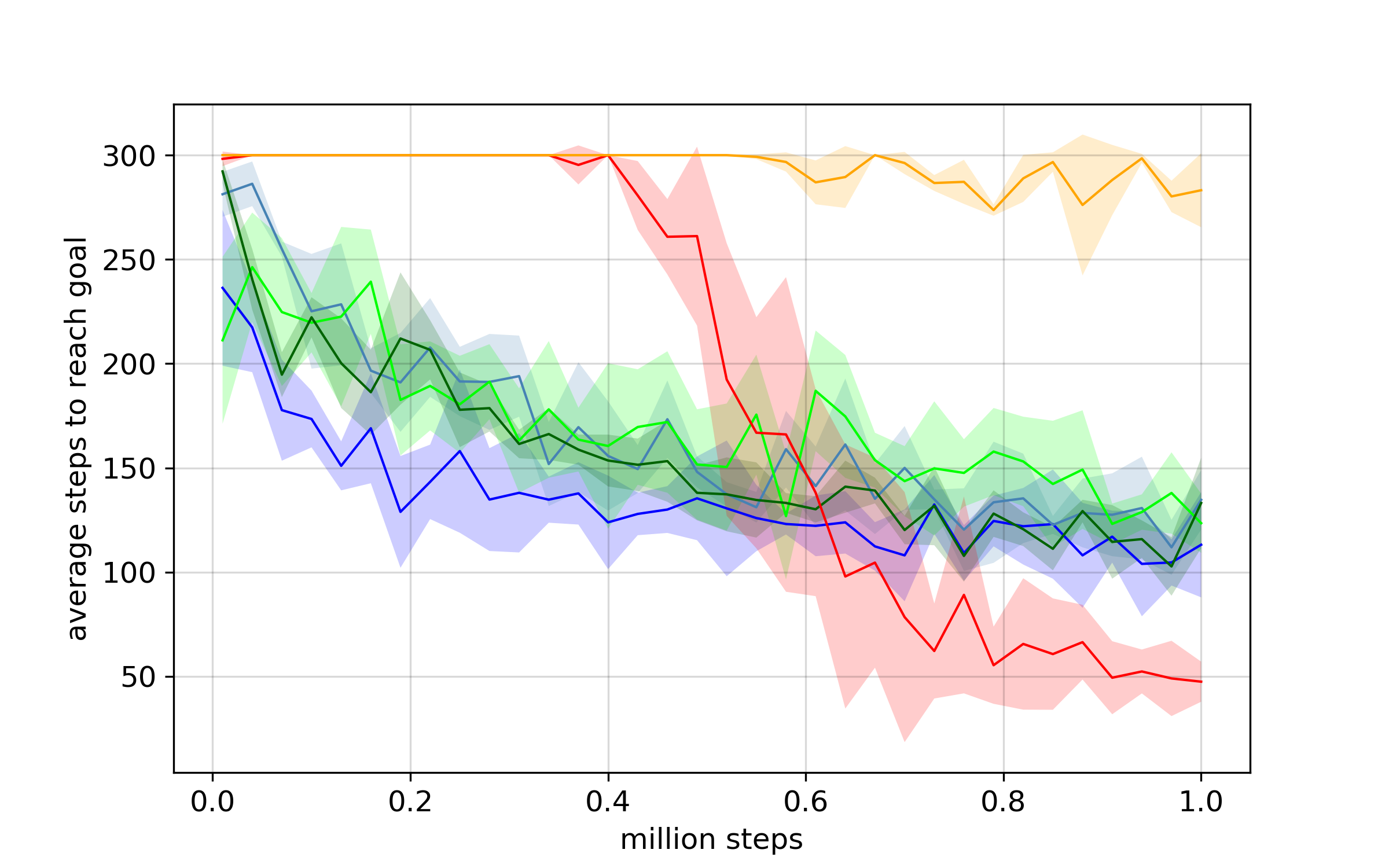}
    		\subcaption{Average number of steps to reach goal (lower is better)}\label{fig:results_steps}
    	\end{minipage}
    	\begin{minipage}[]{0.12\linewidth}
    		\centering
    		\includegraphics[width=1.1\columnwidth]{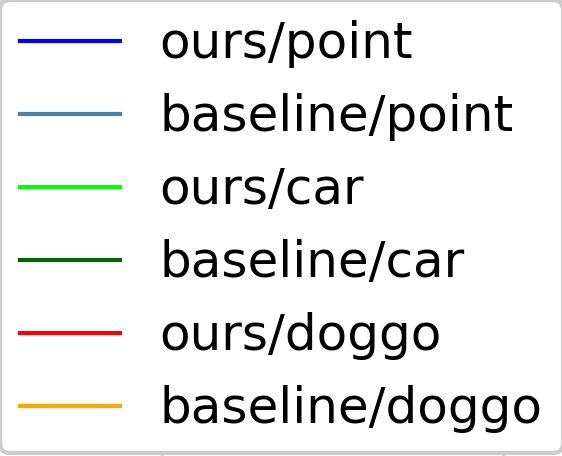}
    	\end{minipage}
    	\caption{Training curves on each robot type. The solid lines represent average returns over five instances with different random seeds. The shaded region represents the standard deviation of the five instances. Our approach outperforms baseline on both sample efficiency and final performance.}
    	\label{fig:sample_efficiency}
    \end{figure*}

    We test the proposed method for sample efficiency as we believe the path-conditioned RL makes the agent explore efficiently. 
    To verify this, we compare the proposed method to an RL agent without waypoints, which we define as baseline. 
    Since the baseline does not use waypoints, it requires a reward signal instead of moving towards goal location along with waypoints. In order to do that, we add a goal distance penalty to encourages the agent to reach the goal, and also add the relative location of the goal $\bm{x}_t^{\rm rel} = \bm{x}_t - \bm{x}_{\rm goal}$ to an input to the RL agent.
    Therefore, the reward function of the baseline will be:
	\begin{equation}
	    r_{\rm baseline}(s_t, a_t) = f(s_t, a_t) + w_5 d_{\rm goal},
    	\label{eq:reward_baseline}
	\end{equation}
    where $f(s_t, a_t)$ is the same reward as defined in \eref{eq:reward_rl}, and the second term is distance penalty to the goal location computed by $d_{\rm goal} = \|\bm{x}_t - \bm{x}_{\rm goal}\|$.
    For a fair comparison, we evaluate both agents on this reward function. Both agents are trained with the SAC algorithm described earlier in Section~\ref{subsec:sac}.

    \Fref{fig:sample_efficiency} shows the training curves of resulting episodic returns.
    It suggests that the use of waypoints improves convergence performance with respect to the training without waypoints.
    Looking at the performance gap between {\point}   and \doggo, we can see that decoupling the planning and control improves sample efficiency especially for high dimensional dynamical system.


\subsection{Generalization to Novel Environments}\label{subsec:results_generalization}

\begin{figure}[t]
    	\begin{minipage}[]{0.49\linewidth}
    		\centering
    		\includegraphics[width=\columnwidth]{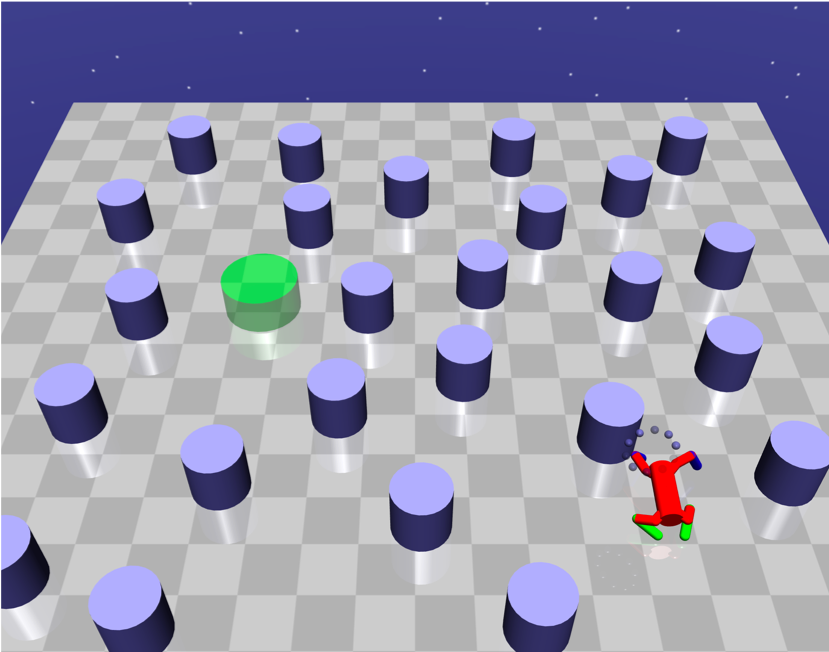}
     		\subcaption{\textit{pillar} $(3, 3, 25)$}
    	\end{minipage}
    	\begin{minipage}[]{0.49\linewidth}
    		\centering
    		\includegraphics[width=\columnwidth]{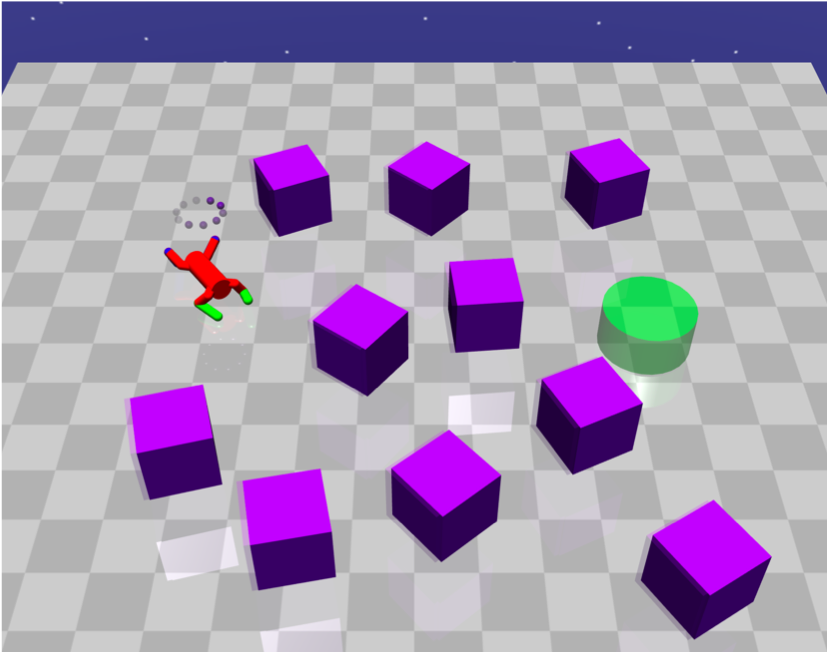}
    		\subcaption{\textit{gremlin}}
    	\end{minipage}
    	\begin{minipage}[]{0.49\linewidth}
    		\centering
    		\includegraphics[width=\columnwidth]{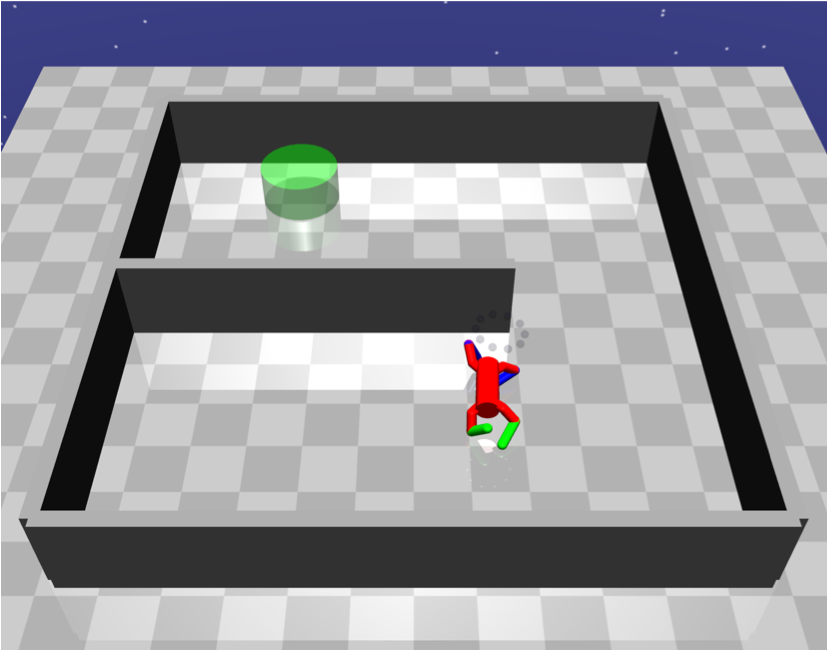}
     		\subcaption{\textit{two-room}}
    	\end{minipage}
    	\begin{minipage}[]{0.49\linewidth}
    		\centering
    		\includegraphics[width=\columnwidth]{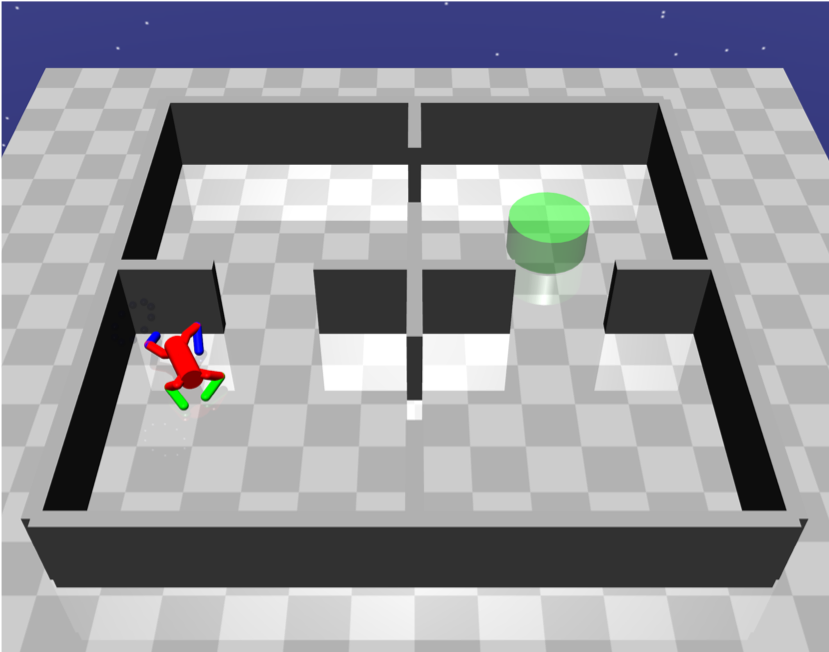}
    		\subcaption{\textit{four-room}}
    	\end{minipage}
    	\caption{Novel environments to verify generalization capability of proposed methods.}
    	\label{fig:generalization_env}
    \end{figure}

    Next, we evaluate the generalization of our method with respect to novel environments.
    As for environment, we prepared five different types of environment as depicted in \fref{fig:generalization_env}.
    We first see whether trained agent can generalize to different size of a field and number of obstacles by preparing \textit{pillar} $(W\times H\times N)$ environment, where $W$ and $H$ is the width and height of the field, and $N$ is number of obstacles. Note that the agent is trained with a \textit{pillar} $(2, 2, 10)$ environment.
    Next, we evaluate whether the agent can generalize to quickly moving obstacles in \textit{gremlin} environment, where boxes move circularly~\cite{Ray2019safetygym}.
    Finally, we evaluate the agent on simple maze environments: \textit{two-room} and \textit{four-room}, where the agent must take a detour to reach the goal location because the rooms are divided by walls, .

    We compared our method with baseline agent, which is the same with the one trained in the previous experiment in \sref{subsec:vs_baseline}.
    The robot type we used is \doggo, which has the most complex dynamical system provided by Safety Gym.
    For a fair comparison, we used a converged model for both ours and baseline trained with three million time steps.
    Since the distribution of input image to the waypoints generator is different from trained settings in \sref{subsec:results_way_points_gen}, we also evaluate the agent trained by our method with fine-tuned waypoints generators. Note that only waypoints model is fine-tuned, and we do not fine-tune the RL agents.

    Table \ref{table:goal_param_success_rate} shows the result of the experiments.
    Overall, our method generalizes to novel environments compared to baseline.
    Since our method decouples planning and control, the fine-tuned waypoints improves the performance in most novel environments by mitigating the inaccuraccy of the waypoints.
    Thus, we verified our method improves generalization capability.
	\begin{table*}[tbp]
		\caption{Performance on generalization task in simulation. The results are averaged over $200$ episodes with random agent, goal, obstacles locations.
		         The bold number indicates the best results among three different methods. In the ``Ours Fine-tuned'', we used the fine-tuned waypoints generator to improve the quality of waypoints. Note we do not retrain the RL agent.}
		\label{table:goal_param_success_rate}
		\begin{center}
			\begin{tabular}{ccccccccccc} \toprule
			   & \multicolumn{3}{c}{Goal reach rate} & & \multicolumn{3}{c}{Steps to reach goal}\\
                          \cmidrule{2-4}
                          \cmidrule{6-8}
			     Environment & Baseline & Ours  & Ours Fine-tuned & & Baseline & Ours & Ours Fine-tuned \\ \midrule
			     \textit{gremlin} & 0.67 & 0.99 & $\mathbf{1.00}$ & & 131.3 & $\mathbf{33.7}$ & 33.8 \\
			     \textit{pillar} $(3, 3, 25)$ & $0.78$ & $0.96$ & $\mathbf{0.97}$ & & $152.1$ & $66.3$ & $\mathbf{59.6}$ \\
			     \textit{pillar} $(4, 4, 40)$ & 0.63 & $\mathbf{0.97}$ & 0.95 & & 273.7 & $\mathbf{76.9}$ & 94.1 \\
			     \textit{two-room} & 0.37 & 0.59 & $\mathbf{0.99}$ & & 397.1 & 268.4 & $\mathbf{59.0}$ \\
			     \textit{four-room} & 0.67 & 0.9 & $\mathbf{0.98}$ & & 231.7 & 64.6 & $\mathbf{50.6}$  \\
			     \bottomrule
			\end{tabular}
		\end{center}
	\end{table*}



\subsection{Online Replanning}
\begin{figure}[tpb]
		\centering
		\includegraphics[width=\columnwidth]{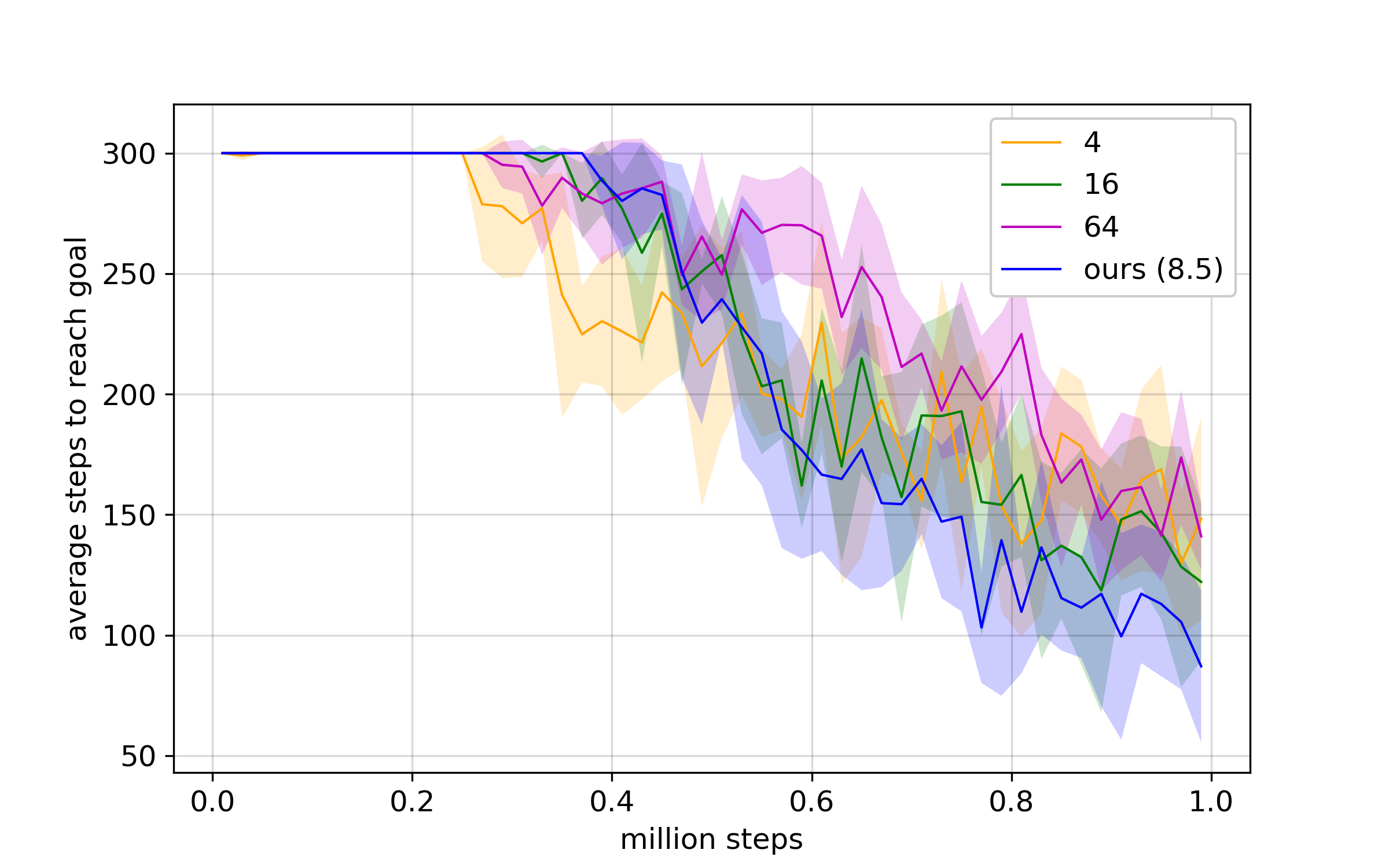}
		\caption{Training curves comparison among different frequencies to update waypoints. The training is done by \doggo.}
		\label{fig:control_freq}
	\end{figure}

    In order to verify the performance of our online replanning strategy described in \sref{subsec:online-replanning}, we conducted a comparative study to evaluate the performance among different frequencies to replan waypoints.
    \Fref{fig:control_freq} shows the results of training curves of average steps to reach goal over different frequencies to update waypoints.
    This results demonstrate that if the replanning frequency is too slow, then the agent does not learn well because of the inaccuracy of the waypoints from which the agent acquires rewards.
    Our proposed method provides more informative reward signals to an agent by replanning the waypoints while the update frequency of $8.5$ is better than $4$.

    \section{Conclusion}
In this paper, we present a sample efficient and generalizable method for navigation task for robots with complex dynamics in obstacle-cluttered environments.
The method combines traditional path planning algorithm, supervised learning, and reinforcement learning.
The SL learns to produce the output of the traditional path planning algorithm as a form of waypoints, and RL takes the waypoints to generate primitive actions to control robots.
In the experiments performed in the Safety Gym suite, we demonstrated that the proposed method 1) learns more sample efficiently than RL without waypoints, 2) generalize well to unknown environments, 3) can adapt to moving obstacles in the environment while ensuring safety of the agent.

In our future work, we plan to extend this method to real application for mobile robots and robotic arms. 

    \bibliographystyle{unsrt}  
    \bibliography{references}

    \section{Appendix}
\subsection{Experiments Details}

\subsubsection{Simulation Setup}
We evaluate our methods on recently proposed Safety Gym environments~\cite{Ray2019safetygym}.
As for obstacles, we located $N$ number of ``pillars'', where $N=10$ for \sref{subsec:vs_baseline} and $N = \{25, 40\}$ for \sref{subsec:results_generalization}, which are immovable objects and the agent gets penalty when it collides with them.
The location of an agent, a goal, and obstacles are randomly sampled when an episode starts, and the goal and obstacles are fixed during an episode.
The field size $(W \times H)$ is $W = H = 2$ [m] as default, and we change the size only when we conduct generalization experiments as described in \sref{subsec:results_generalization}.

\subsubsection{Terminal Conditions}
For the maximum steps for one of terminal conditions, we used $150 \times W$, where $W$ is the width of the field. When the environment is \textit{two-room} or \textit{four-room}, we doubled the episode length because the agent needs to take a detour to reach a goal location.
Also, the episode terminates when the distance between the goal and the agent is less than $0.3$ [m], as implemented in Safety Gym.
The last condition is whether the agent exceeds the limited region $W\times H$, which are defined above.

\subsubsection{Waypoints Generator}
As for the distance condition of updating the waypoints $d_{\rm update}$, which is defined in \sref{subsec:online-replanning}, we use $d_{\rm update}=0.3$ for all experiments.

\subsubsection{Rewards}
Table.~\ref{tab:rewards} summarizes the coefficients of reward terms defined in Eq.~\eqref{eq:reward_way_points},\eqref{eq:reward_rl},\eqref{eq:reward_baseline}.

\begin{table}[tbp]
    \centering
    \caption{Coefficients of each reward term $w_i$ used in our experiments.}
    \begin{tabular}{crl} \toprule
        Term & Value & Description \\ \midrule
         $w_1 d_{\rm path}$ & $-0.1$ & Distance penalty to the closest reference path \\
         $w_2 n_{\rm progress}$ & $0.5$ & Distance reward going toward the goal \\
         $w_3 \mathbb{I}_{\rm collision}$ & $-1.0$ & Obstacle collision penalty \\ 
         $w_4 \mathbb{I}_{\rm goal}$ & $1.0$ & Goal reach reward \\
         $w_5 d_{\rm goal}$ & $-1.0$ & Distance penalty from the goal \\
         \bottomrule
    \end{tabular}
    \label{tab:rewards}
\end{table}

\subsection{Training Details}
\subsubsection{Waypoints Generator}
\Tref{tab:cnn_architecture} summarizes the architecture of the CNN.
This architecture is almost same with previous works for Atari environments~\cite{mnih2015}.

The dataset of the waypoints generator is generated using A* search on each environment type by discretizing $(W \times H)$ field by $0.1$ [m], and collision detection is done by MuJoCo's function~\cite{todorov2012mujoco}.
To mitigate the jerky path caused by the planning on discretized space, we short-cut the produced vertices with randomly selected two vertices, and replace them with new vertices, that connects the two vertices with fixed distance, if those new vertices do not collide with obstacles.

\begin{table}[tbp]
    \centering
    \caption{The architecture of the waypoints generator.}
    \begin{tabular}{cccc}
        \toprule
        Shape & Layer size & Stride & Type \\ \midrule
        $64 \times 64 \times 6$ & - & - & GOSELO-style input \\
        $32 \times 32 \times 32$ & $3 \times 3 \times 32$ & 2 & Convolution \\
        $16 \times 16 \times 64$ & $3 \times 3 \times 64$ & 2 & Convolution \\
        $8 \times 8 \times 64$ & $3 \times 3 \times 64$ & 2 & Convolution \\
        $64$ & - & - & Global Average Pooling \\
        $256$ & - & 256 & Fully Connected \\
        $2 \times n_{way-points}$ & - & - & Output \\
        \bottomrule
    \end{tabular}
    \label{tab:cnn_architecture}
\end{table}

\subsubsection{Reinforcement Learning}
We used SAC agent for an RL agent over all experiments.
The network architecture and hyper-parameters of SAC agent is strictly same with original paper~\cite{haarnoja2018soft}.

\end{document}